\documentclass[10pt,twocolumn,letterpaper]{article}
\usepackage[textwidth=6.875in,textheight=8.875in,left=0.8125in,top=1in,columnsep=0.3125in]{geometry}
\usepackage{times}
\usepackage[dvipsnames]{xcolor}
\usepackage{graphicx,amsmath,amssymb,booktabs,etoolbox,cuted}
\usepackage[numbers,sort&compress]{natbib}
\usepackage[hyphens]{url}
\usepackage[format=plain,labelformat=simple,labelsep=period,font=small,compatibility=false]{caption}
\usepackage[font=footnotesize,skip=3pt,subrefformat=parens]{subcaption}

\usepackage[shortlabels,inline]{enumitem}
\setlist[itemize]{noitemsep,leftmargin=*,topsep=0em}
\setlist[enumerate]{noitemsep,leftmargin=*,topsep=0em}
\font\subsectionheadingfont=ptmb scaled 1100
\font\subsubsectionheadingfont=ptmb scaled 1000
\makeatletter
\def\preprintsection{\@startsection{section}{1}{\z@}{-10pt plus -2pt minus -2pt}{7pt}{\large\bf}}
\def\preprintssect#1{\preprintsection*{#1}}
\def\preprintsect#1{\preprintsection{\texorpdfstring{\hskip -1em.~}{}#1}}
\def\section{\@ifstar\preprintssect\preprintsect}
\def\preprintsubsection{\@startsection{subsection}{2}{\z@}{-8pt plus -2pt minus -2pt}{5pt}{\subsectionheadingfont}}
\def\preprintssubsect#1{\preprintsubsection*{#1}}
\def\preprintsubsect#1{\preprintsubsection{\texorpdfstring{\hskip -1em.~}{}#1}}
\def\subsection{\@ifstar\preprintssubsect\preprintsubsect}
\def\preprintsubsubsection{\@startsection{subsubsection}{3}{\z@}{-6pt plus -2pt minus -2pt}{3pt}{\subsubsectionheadingfont}}
\def\preprintssubsubsect#1{\preprintsubsubsection*{#1}}
\def\preprintsubsubsect#1{\preprintsubsubsection{\texorpdfstring{\hskip -1em.~}{}#1}}
\def\subsubsection{\@ifstar\preprintssubsubsect\preprintsubsubsect}
\def\@maketitle{%
  \newpage\null\vskip .375in
  \begin{center}
    {\Large\bf\@title\par}
    \vspace*{12pt}{%
      \large\lineskip .5em
      \begin{tabular}[t]{c}\@author\end{tabular}\par
    }
    \vskip .5em
  \end{center}
}
\renewenvironment{abstract}{\centerline{\large\bf Abstract}\vspace*{12pt}\noindent\it\ignorespaces}{\vspace*{12pt}}
\makeatother
\date{}

\usepackage{cuted} %
\usepackage{caption} %

\usepackage{bm}
\usepackage{amsthm}
\usepackage{multirow}
\usepackage{colortbl}
\usepackage{xspace}
\usepackage{microtype}
\newtheoremstyle{compact}{3pt}{3pt}{\itshape}{}{\bfseries}{.}{0.5em}{}
\theoremstyle{compact}
\newtheorem{proposition}{Proposition}

\AtEndPreamble{%
  \crefname{proposition}{Prop.}{Props.}%
  \Crefname{proposition}{Proposition}{Propositions}%
  \crefname{lemma}{Lemma}{Lemmas}%
  \Crefname{lemma}{Lemma}{Lemmas}%
}

\newcommand{\E}{\mathbb{E}}
\newcommand{\KS}{\mathrm{KS}}
\newcommand{\KShat}{\widehat{\mathrm{KS}}}
\newcommand{\Var}{V}
\newcommand{\mQ}{m_{Q}}
\newcommand{\mP}{m_{P}}
\newcommand{\abar}{\bar{\alpha}}
\newcommand{\up}{\ensuremath{\uparrow}}

\makeatletter
\renewcommand\paragraph{\@startsection{paragraph}{4}{\z@}{0.9ex \@plus .1ex \@minus .1ex}{-1em}{\normalfont\normalsize\bfseries}}
\makeatother

\makeatletter
\newcommand{\paperdisplayskips}{%
  \setlength{\abovedisplayskip}{4pt \@plus 1pt \@minus 1pt}%
  \setlength{\belowdisplayskip}{4pt \@plus 1pt \@minus 1pt}%
  \setlength{\abovedisplayshortskip}{1pt \@plus 1pt}%
  \setlength{\belowdisplayshortskip}{3pt \@plus 1pt \@minus 1pt}}
\g@addto@macro\normalsize{\paperdisplayskips}
\makeatother
\paperdisplayskips

\definecolor{linkblue}{rgb}{0.12,0.30,0.52}
\usepackage[breaklinks,colorlinks,allcolors=linkblue]{hyperref}
\usepackage[capitalize]{cleveref}
\crefname{section}{Sec.}{Secs.}\Crefname{section}{Section}{Sections}
\crefname{table}{Tab.}{Tabs.}\Crefname{table}{Table}{Tables}
\title{What One Sample Cannot See:\\ Why Fewer Denoising Steps Can Score Higher on Physics Benchmarks}
\author{Yutao Li, Chen Wang\textsuperscript{*}\\[2pt]
Harbin Institute of Technology\\[2pt]
{\small \textsuperscript{*}Corresponding author: Chen Wang (\href{mailto:chen.wang@hit.edu.cn}{chen.wang@hit.edu.cn})}}
\hypersetup{pdftitle={What One Sample Cannot See: Why Fewer Denoising Steps Can Score Higher on Physics Benchmarks},pdfauthor={Yutao Li, Chen Wang},pdfsubject={Preprint}}

\usepackage{balance}
\usepackage{fancyhdr}
\fancypagestyle{plain}{%
  \fancyhf{}%
  \fancyhead[L]{\footnotesize\itshape\color{gray}Preprint}%
  \fancyfoot[C]{\thepage}%
}

\makeatletter
\g@addto@macro\normalsize{\setlength{\baselineskip}{12.15pt}}
\makeatother
\normalsize

\begin{document}

\maketitle
\begin{strip}
\centering
\includegraphics[width=\textwidth]{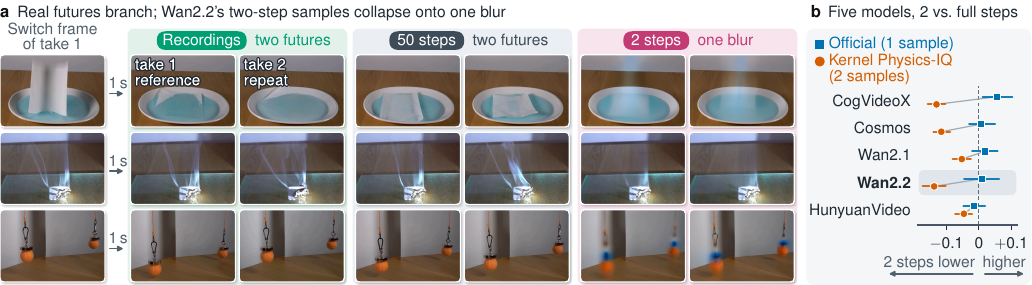}
\captionsetup{hypcap=false}
\captionof{figure}{\textbf{One sample cannot see dispersion.} (a) Two takes of one setup and two 50-step Wan2.2 samples from the switch frame of take 1 branch into distinct futures, whereas two-step samples collapse onto one blur; official scores compare each sample with take 1 alone. (b) Two-step sampling lowers Kernel Physics-IQ for all five models; official scores rise on CogVideoX and show no detected loss elsewhere (95\% CIs; Wan2.2 shaded).}
\captionsetup{hypcap=true}
\label{m:fig:teaser}%
\end{strip}
\begin{abstract}
When a storm can head north or south, a forecast down the middle may score better under squared error than either plausible path.
The same paradox appears when video generators are tested as world models: fewer denoising steps can raise physics scores even as videos blur or stop moving early.
Why?
Physics-IQ compares each generated video with one recording, although a scene can have multiple valid futures.
An exact decomposition shows its scores can reward reduced variation among generated futures, even when their distribution becomes less accurate.
We introduce Kernel Physics-IQ to remove this incentive in expectation, needing only two independent samples per condition.
Two-step sampling lowers Kernel Physics-IQ in all five image-to-video models.
For CogVideoX, the official score rises by 0.055 while ours falls by 0.131.
Changing the sampling schedule reveals further differences official scores miss.
In controlled physical systems with known outcome probabilities, our metric favors the true distribution over collapsed predictions.
In short, a world model should be judged not only by how closely one video matches what happened, but by how faithfully its predictions represent what could have happened.%
\end{abstract}
\section{Introduction}
\label{m:sec:intro}

\begin{figure*}[t]
\centering
\includegraphics[width=\textwidth]{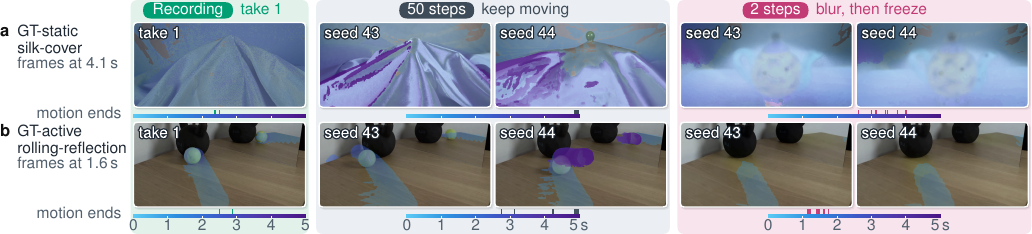}
\caption{\textbf{Two-step samples blur, agree across seeds and freeze early.} Wan2.2 on silk-cover (a) and rolling-reflection (b), scenes whose 50-step samples keep objects intact: each frame carries its 0--5\,s motion mask, colored by when each pixel last moves; ticks mark when each take or seed stops. In silk-cover (takes settled by 2.5\,s), all 50-step seeds still move at 5\,s; two-step seeds settle on one blurred teapot by 4\,s. In rolling-reflection (takes stop by 2.9\,s), two-step seeds smear the ball and stop by 1.75\,s; 50-step seeds roll until 2.75--5\,s.}
\label{m:fig:qual}
\end{figure*}

Video generators are increasingly evaluated as world models that predict how objects move, collide and interact~\cite{physicsiq,cosmos3,wan}.
Yet few-step sampling exposes a striking leaderboard puzzle.
PhaseLock~\cite{phaselock} observed that CogVideoX~\cite{cogvideox} scores higher on Physics-IQ~\cite{physicsiq} after 2 denoising steps than after 50.
Measuring reduced low-frequency phase coherence across those steps, PhaseLock guided sampling with a two-step prior to raise official scores across three models.
In the generated videos themselves, however, two-step outputs blur moving objects, agree across seeds and stop moving early (\cref{m:fig:teaser}a, \cref{m:fig:qual}).
Judge-based evaluation points the opposite way: on PhyWorldBench~\cite{phyworldbench}, which rates physical realism without a reference recording, Open-Sora improves with more denoising steps.

Why can visibly degraded videos score higher on a reference-based physics benchmark?
Film the same experiment twice, and small unseen differences send the two takes down distinct trajectories (\cref{m:fig:teaser}a).
A calibrated world model represents this distribution of possible futures, drawing one plausible branch per seed instead of aiming at a single outcome.
Physics video benchmarks, however, usually compare one generated video with one real recording~\cite{physicsiq,physicsiqverified}: Physics-IQ releases two takes of each scene, yet scores every sample against take 1 alone and uses take 2 only for normalization.
When nature takes another branch, a faithful draw lands far away, while a middle-ground prediction stays closer on average.
Weather models trained with MSE likewise ``hedge their bets'' by predicting a blurry mean state with low RMSE~\cite{weatherbench2}.
Meanwhile, a large human study asks whether metrics should target ``a specific observed outcome'' or ``the broader space of physically plausible outcomes''~\cite{rapidata}.
A basic question remains open for reference benchmarks: how much of the few-step gain reflects a closer predictive distribution, and how much comes from score-induced hedging?

Our key insight is that single-sample distance splits exactly.
Consider any negative-type distance, meaning one embeddable as squared Euclidean distance in a feature space~\cite{schoenberg1938}.
For any such distance, the expected distance between one sample and one recording equals a proper kernel score plus half the model's self-dispersion, an identity from forecast verification~\cite{gneiting2007,ferro2014}.
We show that every Physics-IQ component belongs to this class, namely pixel MSE and the three thresholded motion-mask IoUs with their empty-mask convention, as does PhaseLock's phase-coherence diagnostic on a fixed frequency support.
Because half self-dispersion depends only on pairs of model outputs and never sees the recording, shrinking dispersion improves the single-sample score whenever it saves more than the kernel score loses.
We call the gain bought this way the \emph{dispersion premium}.
A branching pendulum gives this premium physical geometry.
When two valid futures swing left and right, their mean position sits on the chord midpoint inside the arc like a shorter pendulum, saving under squared distance exactly the drop in squared length; in pixels it is a double exposure.
On binary masks, where superimposing paths fails, uncertain motion rewards standing still.

The identity and kernel-score properness are classical; our contribution is to make them operational for a physics benchmark built from thresholded masks and an empty-mask convention, and to show that this changes conclusions about real world models (\cref{m:fig:teaser}).
First, we show that any official Physics-IQ difference on a fixed scene set splits with the official weights into a kernel part and a dispersion premium.
Second, we introduce \emph{Kernel Physics-IQ}, an unbiased two-sample score normalized by 52 non-duplicate take pairs that replaces pixel MSE (whose kernel sees only the pixel mean) with full-video RMS.
Third, across five image-to-video world models, we show that default two-step sampling lowers Kernel Physics-IQ for every model (\cref{m:fig:teaser}b).
On CogVideoX, its $+0.055$ official gain reflects a 0.200 dispersion premium hiding a 0.145 kernel loss, as cross-seed overlap jumps far faster than overlap with the recording.
At fixed weights across CogVideoX, Cosmos and Wan2.1, shifting only the two-step node raises Kernel Physics-IQ with no detected official difference (\cref{m:fig:steps}a).
Finally, in synthetic worlds with known distributions, official scoring ties faithful sampling with one-outcome predictors and mixtures with wrong probabilities, whereas Kernel Physics-IQ ranks the true distribution highest.
Adding one sample per condition removes the incentive to hedge in expectation, so we recommend reporting Kernel Physics-IQ alongside sample dispersion.
\section{Related Work}
\label{m:sec:related}

\paragraph{Physics video benchmarks and stochastic evaluation.}
Physics-IQ~\cite{physicsiq} grounds evaluation in real experiments by comparing one continuation against a recording via three motion-mask IoUs and pixel MSE, filming a second take as a physical-variance reference.
Our frozen constants build on that two-take design, while Physics-IQ Verified~\cite{physicsiqverified} revises prompts and references with per-scene truncation.
Complementary benchmarks ask human or model judges whether physics is obeyed~\cite{videophy,phygenbench,worldmodelbench,phyworldbench,phyground} or measure physical quantities~\cite{physeval}, with PhyGround noting that pixel references assume a unique future.
For multi-sample evaluation, CaliBench~\cite{calibench} tests calibration on discrete outcomes with known distributions, and probabilistic segmentation uses generalized energy distances~\cite{kohl2018,zepf2024}. Probabilistic video forecasts report per-pixel CRPS and spread--skill ratios~\cite{yang2023rvd,cachay2023dyffusion}, whereas stochastic video and trajectory prediction often score the best of several samples~\cite{lee2018savp,gupta2018socialgan}, rewarding wider dispersion instead of calibration.
We keep real video references and the two-take design, removing the dispersion premium with two samples scored on whole clips.

\paragraph{Proper scoring rules and ensemble verification.}
Forecast verification formalized proper scoring rules and score-induced hedging~\cite{murphy1967,murphy1973,gneiting2007}, showing why an ensemble mean halves one member's squared error~\cite{leith1974} and how fair scores remove the under-dispersion preference of finite ensembles~\cite{ferro2014}.
Sample-only losses require at least $k$ draws to identify degree-$k$ distributional functionals~\cite{kimpara2023}, and negative-type kernel scores measure squared MMD~\cite{sejdinovic2013}.
Machine-learning weather models therefore report CRPS, spread--skill ratios and spectra~\cite{weatherbench2,gencast,subich2025}, alongside kernel-score decompositions, spatio-temporal signature kernels and single-target proper scores~\cite{gruber2024,dodson2025,tonks2026}.
We bring this framework to physics video benchmarks across thresholded motion masks, empty-mask conventions and phase coherence.

\paragraph{Few-step sampling, guidance and sample shrinkage.}
PhaseLock~\cite{phaselock} studies the few-step phenomenon from the generation side, measuring low-frequency phase-coherence decay across steps and guiding sampling with a two-step prior.
Diffusion theory explains how coarse discretization reshapes sample distributions: MSE-trained denoisers return conditional means~\cite{debortoli2025}, Gaussian trajectories follow exact solutions~\cite{pierret2025} and cosine-product shrinkage~\cite{ren2025}, and critical windows govern when trajectories commit to outcomes~\cite{biroli2024,raya2023,li2024critical}.
Likewise, distillation and strong guidance narrow generated distributions~\cite{gandikota2026,sadat2024cads}, while perception--distortion theory bounds the lowest MSE at perfect perception to at most twice the MMSE~\cite{blau2018,freirich2021}.
Whereas these works study how sampling alters generated videos, we study how those shifts move single-sample scores.
\section{Method}
\label{m:sec:theory}
Evaluating a stochastic physical scene from one generated continuation conflates distributional accuracy with sample shrinkage. Separating the two begins by asking what one sample can and cannot identify.

\subsection{Single-sample risk rewards shrinkage}
\label{m:sec:linear}\label{m:sec:split}
\begin{figure}[t]
\centering
\includegraphics[width=\linewidth]{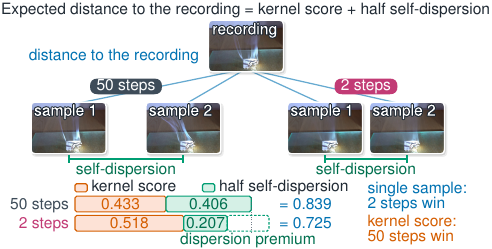}
\caption{\textbf{What one sample cannot see.} A sample is scored by its distance to the recording (blue), which on average is the kernel score plus half the self-dispersion (green). Two-step samples disperse less: the dispersion premium they collect (dashed) outweighs their kernel loss, so single-sample scoring prefers them. A second sample measures the self-dispersion, which Kernel Physics-IQ subtracts. Frames: paper-smoke 2\,s after the switch frame (Wan2.2, two seeds). Brackets (root-mean-square) and bars: Wan2.2 ST IoU distances on whole clips, mean over 66 scenes.}
\label{m:fig:method}
\end{figure}
Judging a stochastic world model from one generated video is like reading depth with one eye closed.
A single trajectory shows where one draw landed, yet gives no view of the distribution behind it.
For a condition $c$, let $r_c(x)=\E_{Y\sim P_c}\,d(x,Y)$ be the expected distance from video $x$ to the real future $Y\sim P_c$.
Evaluating one draw from the predictive distribution $Q_c$ gives the expected score
\begin{equation}
R(Q_c;P_c)=\int r_c(x)\,\mathrm{d}Q_c(x),
\label{m:eq:linear}
\end{equation}
which is linear in $Q_c$.
Any linear functional of a distribution attains its minimum at a point mass.
If $r_c$ achieves its infimum at $x^*$, the deterministic forecast $\delta_{x^*}$ is optimal, and a calibrated sampler scores strictly worse whenever $r_c$ is not $P_c$-almost surely constant.
This is the Bayes act of decision theory~\cite{gneiting2011} and the video-benchmark form of a classic verification principle: one sample cannot identify a distribution~\cite{ferro2014,kimpara2023}.
The log score evaluates a full density from one outcome, whereas black-box generators expose samples instead of likelihoods.

A reference-based benchmark therefore favors a compromise close to plausible futures (the conditional mean under pixel MSE), whereas a judge-based benchmark favors the judge's preferred trajectory.
For an exact denoiser, one step from pure noise returns the conditional mean and coarse steps contract Gaussian modes toward it (\cref{m:sec:ideal}); in most real models we test, fewer steps shrink dispersion.
The two kinds of benchmark can thus rank step counts in opposite directions while neither rewards calibration.

For Physics-IQ this shrinkage incentive splits in closed form because its components are negative-type distances (\cref{m:sec:components}), embeddable as squared Euclidean distances $d(x,y)=\lVert\phi(x)-\phi(y)\rVert^2$~\cite{schoenberg1938}.
With $\mQ=\E_Q\phi(X)$ as the model's kernel mean embedding, any distribution $Q$ and observation $y$ satisfy
\begin{equation}
\E_{X\sim Q}\,d(X,y)=\underbrace{\lVert \mQ-\phi(y)\rVert^2}_{\KS_d(Q,y)}+\underbrace{\tfrac12\,\E\,d(X,X')}_{\Var_d(Q)}.
\label{m:eq:split}
\end{equation}
\indent The left side is what a single-sample benchmark sees; on the right, $\KS_d$ is the proper kernel score~\cite{gneiting2007}, and $\Var_d$, half the self-dispersion of independent draws $X,X'\sim Q$, never sees the recording.
Averaging over $Y\sim P$ gives
\begin{equation}
R(Q;P)=c+\lVert \mQ-\mP\rVert^2+\Var(Q),\;\; c=\tfrac12\,\E\,d(Y,Y').
\label{m:eq:risk}
\end{equation}
\indent Single-sample risk sums outcome uncertainty $c$, distributional discrepancy $\lVert \mQ-\mP\rVert^2$ (MMD$^2$~\cite{sejdinovic2013}), and half self-dispersion $\Var(Q)$ paid by a calibrated model.

\paragraph{Dispersion premium.}
Comparing a setting $\theta$ against a reference setting splits the change in single-sample risk into $\Delta\KS+\Delta\Var$.
We call $\Var(Q_{\mathrm{ref}})-\Var(Q_\theta)$ the \emph{dispersion premium}: the single-sample gain collected by shrinking dispersion.
On a fixed scene set, the raw official composite is affine in the component means.
Any official difference therefore splits with the official weights into a kernel part and a dispersion premium.
When the premium exceeds the kernel loss, the single-sample score improves while the predictive distribution worsens (\cref{m:fig:method}).
This is hedging in the sense of Murphy and Epstein~\cite{murphy1967}: a one-member ensemble score, optimal at zero dispersion~\cite{ferro2014}.

\Cref{m:eq:risk} yields three further consequences.
First, the kernel score is proper, with expected excess $\lVert \mQ-\mP\rVert^2$.
Second, the single-sample optimum is the realizable video with features closest to $\mP$, which need not be any physical outcome, as $0$ is not an outcome of $\pm1$.
Third, a calibrated model has risk $2c$, matching a second take scored against the first.
The best deterministic output instead lies between $c$ and $2c$ (Leith's ensemble relation under pixel MSE~\cite{leith1974}).
A single-sample score can thus beat the take-2 reference without capturing the distribution of futures, whereas the expected kernel score never goes below $c$.

\subsection{Pixels blur while masks reward stillness}
\label{m:sec:components}
Pixel averaging explains MSE gains, yet path superposition fails on binary masks. How do masks reward shrinkage?

\begin{proposition}\label{m:prop:negtype}
The Jaccard distances of the three mask IoUs, pixel MSE and full-video RMS ($\lVert x-y\rVert^\beta$, $\beta=2$ and 1), and phase coherence ($1-\mathrm{PC}$, a squared chord on the unit circle) are negative-type distances. The convention that two empty masks score 1 is a direct sum with a constant block, and averaging over frames preserves negative type~\cite{gower1986,bouchard2013,ioffe2010}.
\end{proposition}

\noindent Every component therefore obeys \cref{m:eq:split} (on a fixed frequency support for phase coherence, used as a diagnostic outside Kernel Physics-IQ), yet $\phi$ induces two distinct ways to collect the dispersion premium.
Under pixel MSE, the optimal output is the conditional mean.
Under phase coherence, a template shifted by a random delay with mean phasor length $R>0$ gives expected coherence $R^2$ for a sample versus $R$ for the mean video.
The mean video gains $R(1-R)$ on the single-sample score while losing $(1-R)^2/2$ on the kernel score.
For binary motion masks, however, pixel averaging fails as a hedge.
Superimposing diverging trajectories leaves a faint cross-fade whose thresholded mask still fires along both paths, losing the empty-mask credit and diluting overlap with whichever branch occurs.
Under the empty-mask convention, uncertain motion instead rewards standing still:

\begin{proposition}\label{m:prop:still}
Let a frame's real future move with probability $p$, and let two empty masks score 1. Staying still scores $1-p$ in expectation, and an output motion mask $A$ scores $pJ(A)$, where $J(A)$ is its expected IoU with the real motion mask. Staying still is optimal when $p\le1/(1+J^*)$, with $J^*$ the best attainable $J$. For $J^*=0.3$, a model should stay still unless motion is at least 77\% certain. Calibrated sampling scores $(1-p)^2+p^2J$, a weighted average of staying still and moving that never beats the better of the two.
\end{proposition}

\noindent Pixel MSE thus rewards blurring toward the mean, whereas binary masks reward suppressing uncertain motion.
Collapsing to the mean leaves the $\beta=2$ kernel score invariant, whereas collapsing to one outcome $a$ raises it by $\lVert\phi(a)-\mP\rVert^2>0$ unless $\phi(a)=\mP$.
The single-sample score can move either way (tested in \cref{m:sec:known}).

\subsection{Two samples isolate Kernel Physics-IQ}
\label{m:sec:kpiq}
Drawing a second independent sample from the same switch frame opens the other eye without requiring a second recording per scene. Their pairwise distance estimates the self-dispersion term in \cref{m:eq:split}.
From $K\ge2$ samples we estimate each component's kernel score with an unbiased U-statistic, and for two samples per condition
\begin{equation}
\KShat_{d,2}=\tfrac12\big[d(x_1,y)+d(x_2,y)\big]-\tfrac12\,d(x_1,x_2).
\label{m:eq:estimator}
\end{equation}
\indent Turning \cref{m:eq:estimator} into a distribution-level composite closes two loopholes exposed by \cref{m:eq:risk}.
First, because a kernel score identifies only $\mQ$, pixel MSE and phase coherence see only mean pixels or mean phasors.
The $\beta=2$ MSE kernel score cannot separate a pixel-mean blur from truth, while mask kernels capture mask features.
We therefore replace pixel MSE with full-video RMS ($\beta=1$, root-mean-squared pixel difference over the full clip), an energy distance whose kernel score uniquely identifies any video distribution with finite first moment.
Second, we avoid per-sample ratios and clipping~\cite{pic2025,murphy1973}, which distort expectations.
Instead, we apply affine normalization using frozen constants $c_j$ from the 52 non-duplicate take pairs (\cref{m:sec:setup}), preserving properness and unbiasedness:
\begin{equation}
\mathrm{KPIQ}=\frac13\sum_{j\in\{\mathrm{S},\mathrm{ST},\mathrm{W}\}}\frac{1-\overline{\KS}_j}{1-c_j}-\big(\overline{\KS}_{\mathrm{RMS}}-c_{\mathrm{RMS}}\big).
\label{m:eq:kpiq}
\end{equation}
\indent With no truncation, the true distribution scores 1 in expectation when $c_j$ equal the population values of $\tfrac12\E\,d(Y,Y')$.
We pair two readouts throughout: the official-weight split decomposes an official difference, whereas Kernel Physics-IQ, with its own weights and the RMS term, ranks predictive distributions, so its change differs from that kernel part.

\paragraph{Two overlaps.}
For each mask component, let $x$ be the mean IoU of a sample with the recording and $y$ the mean IoU between two samples.
Then single-sample distance is $1-x$, half self-dispersion is $(1-y)/2$, and the kernel score is $(1-2x+y)/2$.
With $w_j=1/\big(3(1-c_j)\big)$, the mask part of \cref{m:eq:kpiq} becomes
\begin{equation}
\sum_{j}w_j\big(1-\overline{\KS}_j\big)=0.695+\sum_{j}w_j\Big(x_j-\frac{y_j}{2}\Big),
\label{m:eq:overlap}
\end{equation}
where $0.695=\tfrac12\sum_j w_j$ is the \emph{no-overlap level}.
The mask terms of the official composite depend only on $x$, whereas the kernel score also accounts for $y$.
Tiny random motion blobs sit near $0.695$, exceeded only when $\bar x>\bar y/2$ for the $w$-weighted means.
Calibrated predictors lie on the diagonal $\bar x=\bar y$; the mask part is 1 at the \emph{take-pair point} $\bar x=\bar y=0.44$, the mean take-pair overlap behind $c_j$.
\section{Experiments}
\label{m:sec:where}

\begin{table}[t]
\centering
\caption{\textbf{Two steps lower Kernel Physics-IQ for every model.} Center-view scores on all 66 scenes (SSR is 1 when calibrated). Bold marks the higher point estimate between default full and two-step runs. HunyuanVideo's SSR falls via sample-mean error without detected net shrinkage. Take 2 is scored as a one-outcome forecast on the 52 non-duplicate pairs.}
\label{m:tab:main}
\footnotesize
\setlength{\tabcolsep}{3.2pt}
\begin{tabular}{@{}lccc@{}}
\toprule
Setting & Official\,\up & Kernel Physics-IQ\,\up & SSR \\
\midrule
CogVideoX, 50 steps & 0.301 & \textbf{0.737} & 0.90 \\
CogVideoX, 2 steps & \textbf{0.356} & 0.607 & 0.29 \\
CogVideoX, LDG (50 steps) & 0.351 & 0.707 & 0.81 \\
\midrule
Cosmos, 35 steps & 0.294 & \textbf{0.725} & 1.09 \\
Cosmos, 2 steps & \textbf{0.301} & 0.609 & 0.51 \\
\midrule
Wan2.1, 50 steps & 0.291 & \textbf{0.746} & 1.21 \\
Wan2.1, 2 steps & \textbf{0.308} & 0.693 & 0.86 \\
Wan2.1, distilled 4 steps & 0.288 & 0.619 & 0.94 \\
\midrule
Wan2.2, 50 steps & 0.396 & \textbf{0.766} & 0.87 \\
Wan2.2, 2 steps & \textbf{0.405} & 0.628 & 0.37 \\
\midrule
HunyuanVideo, 50 steps & \textbf{0.330} & \textbf{0.698} & 0.95 \\
HunyuanVideo, 2 steps & 0.317 & 0.652 & 0.62 \\
\midrule
Take 2 as a prediction & 1.000 & 0.589 & -- \\
\bottomrule
\end{tabular}
\end{table}

Across five open image-to-video models and three inference controls, we test whether few-step gains reflect closer distributions or a dispersion premium, trace how masks collect it, and probe zero dispersion.

\subsection{Benchmarks, frozen constants and models}
\label{m:sec:setup}

\paragraph{Benchmark and composites.}
We evaluate the spatial (S), spatiotemporal (ST) and weighted spatial (W) motion-mask IoUs of Physics-IQ~\cite{physicsiq}, scoring two empty masks as 1, together with pixel MSE, full-video RMS and RGB phase coherence (PC).
The official composite scores every generated video against take 1, whose switch frame conditions all models, averaging the three IoUs normalized by their take-2 means (take 2, a second recording of the same setup, scored against take 1) and subtracting excess MSE over the take-2 mean; its $[0,1]$ clip never binds for any model here, matching the raw composite.
Kernel Physics-IQ applies \cref{m:eq:kpiq} in one aligned RGB space, where 0.01 equals one Physics-IQ point.

\paragraph{Scenes and frozen constants.}
Our main benchmark covers all 66 center-view scenes.
In the released dataset, 14 of the 66 center-view take pairs are byte-identical between takes 1 and 2, as the benchmark README documents~\cite{physicsiqrepo}.
We compute the frozen constants $c_j$ as averages of $d(y,y')/2$ over the 52 non-duplicate pairs, while keeping official take-2 means for the official composite.
Takes start from close but not identical switch frames, so take pairs only approximate repeated draws from $P_c$.
Model comparisons, step sweeps and node shifts use all 66 scenes, while history experiments (\cref{m:sec:history}) use 32 scenes, most chosen for past motion.

\paragraph{Models, controls and inference.}
Our five image-to-video world models anchor on PhaseLock~\cite{phaselock} while spanning DDIM, UniPC and flow-matching Euler solvers.
CogVideoX-5B-I2V~\cite{cogvideox} and Wan2.1-I2V-14B at 480p~\cite{wan} match PhaseLock's two backbones.
Because Wan is among the most widely used open video families, we also evaluate its next open-weight generation, Wan2.2-I2V-A14B~\cite{wan}, pairing dense Wan2.1 with high/low-noise two-expert denoising on the same VAE and UniPC solver.
Cosmos3-Nano-16B~\cite{cosmos3} enables multi-frame history conditioning (\cref{m:sec:history}), and HunyuanVideo-1.5-I2V at 480p~\cite{hunyuanvideo15} contributes an Euler-sampled architecture.
Each model runs with its official scheduler, guidance and resolution using an empty negative prompt, and few-step evaluations alter only step counts.
We also test three inference controls: PhaseLock's two-step prior guidance (LDG) on 50-step CogVideoX, 4-step distilled Wan2.1 (LightX2V) without classifier-free guidance (CFG)~\cite{ho2022cfg,lightx2v}, and 8-step step-distilled HunyuanVideo.
Our official baselines lie within about two points of PhaseLock's released numbers.
Each setting generates seven samples sharing initial noise across step counts, and we call a difference detected when its 95\% paired scene-bootstrap interval excludes zero.
The spread--skill ratio (SSR) compares sample dispersion against the RMSE of the seven-seed ensemble mean with finite-$K$ correction, equaling 1 when calibrated~\cite{weatherbench2}.
We also score that seven-seed average as a sample-mean video.

\subsection{Dispersion premiums drive few-step gains}
\label{m:sec:rank}\label{m:sec:decomp}
\begin{figure}[t]
\centering
\includegraphics[width=\linewidth]{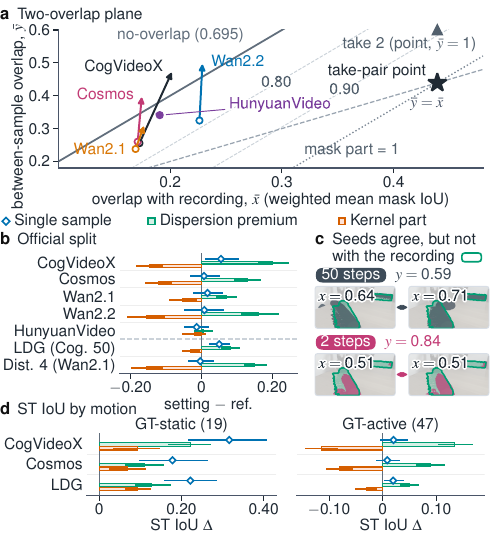}
\caption{\textbf{Fewer steps align seeds far more than they match truth.} (a) Recording ($\bar x$) vs.\ seed overlap ($\bar y$) with \cref{m:eq:overlap} contours and the $0.695$ no-overlap line. (b) Dispersion-premium and kernel-part bars behind net diamonds for the official split. (c) Spatial masks behind $x$ and $y$: Wan2.2 seeds 43 and 44 on rolling-reflection (\cref{m:fig:qual}b) against take 1 (teal). (d) ST IoU splits by scene motion.}
\label{m:fig:decomp}
\end{figure}

Default two-step sampling lowers Kernel Physics-IQ for all five models, with every interval excluding zero (\cref{m:tab:main}, \cref{m:fig:teaser}b).
The decline holds for any two of the seven seeds, under the 66-pair constants, on all three camera views for CogVideoX, Cosmos and Wan2.1, and for Wan2.1 with its official negative prompt.
Yet the official point estimate rises at two steps for four of the five models.
Our official-weight split (\cref{m:fig:decomp}b) explains why: each of those four models collects a detected dispersion premium that exceeds its kernel loss.
On CogVideoX, a 0.200 dispersion premium outweighs a 0.145 kernel loss, producing a detected $+0.055$ official gain while Kernel Physics-IQ drops by 0.131.
All six CogVideoX distances, including phase coherence in RGB, pair smaller self-dispersion with a worse kernel score, as does PhaseLock's latent phase coherence on CogVideoX and Wan2.1.
On Cosmos, Wan2.1 and Wan2.2, the two parts nearly cancel across 66 scenes, while both Wan models retain detected official gains on the 52 non-duplicate scenes.
Two-expert Wan2.2, which sets the highest full-step baselines in \cref{m:tab:main}, collects more than twice the premium of dense Wan2.1 at two steps, so a stronger next-generation backbone still faces this incentive.

HunyuanVideo and the three controls test this mechanism from both directions.
At default two steps, HunyuanVideo shrinks ST dispersion while expanding S dispersion with no detected pixel shift, leaving no detected net shrinkage, premium or official difference.
Conversely, 8-step step-distilled HunyuanVideo cuts pixel MSE dispersion by more than half and immediately collects a detected premium with no detected Kernel Physics-IQ difference.
PhaseLock's guided sampler (LDG) likewise collects a dispersion premium that lifts CogVideoX officially by $+0.050$ while lowering Kernel Physics-IQ by 0.031 (both detected).
In 4-step distilled Wan2.1, the Kernel Physics-IQ drop outpaces shrinkage, leaving no detected official gain.

The two-overlap geometry of \cref{m:eq:overlap} makes this hedging directly visible (\cref{m:fig:decomp}a, c).
Two steps nudge CogVideoX's overlap with the recording ($\bar x$) from 0.17 to 0.20 while nearly doubling its cross-seed overlap ($\bar y$) from 0.26 to 0.47, and the other shrinking models climb steeply upward too.
Full-step CogVideoX, Cosmos and both Wan models stand detectably above the no-overlap level, whereas neither HunyuanVideo nor any default two-step run does.
Few-step sampling raises seed agreement far more than agreement with the recording.

\subsection{Early stillness explains motion-mask gains}
\label{m:sec:static}
Could Kernel Physics-IQ drop simply by penalizing desirable seed consensus while few-step masks trace cleaner trajectories? Separating scenes by real motion argues against that explanation.
CogVideoX's ST gain originates in frames with an empty GT mask while moving frames worsen, so the fewer active frames a scene has, the more it gains on ST IoU.
Splitting scenes in advance by take-1 GT-empty fraction $\ge 0.5$ yields 19 GT-static and 47 GT-active scenes.
CogVideoX concentrates its official few-step gain in GT-static scenes ($+0.190$) with no detected gain in GT-active ones, while its ST kernel score improves on GT-static scenes and worsens on GT-active ones (\cref{m:fig:decomp}d).
Cosmos, LDG and both Wan models share this divide: in GT-active scenes, Kernel Physics-IQ drops for all of them, and Cosmos and both Wan models also lose officially.

Dispersion-free frame counts are consistent with the stillness incentive of \cref{m:prop:still} (\cref{m:fig:qual}).
In GT-static scenes, full-step models move in about half of the frames where the recording is still, and two steps remove much of this added motion.
In GT-active scenes, the four shrinking models miss real motion in 26--42\% of frames at two steps versus 8--10\% at full steps (all differences detected).
A video that never moves, the switch frame repeated over the window, scores within the full-step range officially in GT-static scenes, yet falls far below every full-step model on Kernel Physics-IQ.
Damping take 2 toward its switch frame instead cuts pixel MSE yet lowers the official composite, and pulling 50-step samples toward their mean lowers their official ST IoU.
Two-step mask gains thus come from cross-seed consensus on stillness where full-step samples vary more in their motion.

\begin{figure*}[t]
\centering
\includegraphics[width=\textwidth]{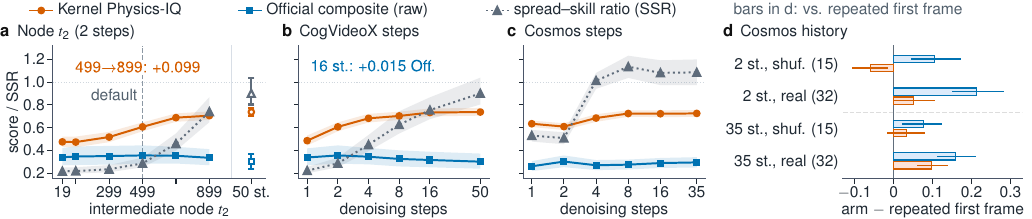}
\caption{\textbf{Fixed-weight interventions expose the dispersion premium.} (a) Sweeping CogVideoX node $t_2$ lifts Kernel Physics-IQ and SSR past 499 with no detected official difference. (b, c) CogVideoX and Cosmos steps. (d) Cosmos history arms ($n=32$, $n=15$ shuffled).}
\label{m:fig:steps}
\end{figure*}

\subsection{Point forecasts expose hedging}
\label{m:sec:ghost}
Pushing shrinkage to zero dispersion tests \cref{m:eq:risk} at its extreme.
On the official composite, averaging seven seeds into a blurry sample-mean video stays within 3.1 points of individual samples across all twelve settings in \cref{m:tab:main} and scores higher in seven, yet Kernel Physics-IQ collapses the same videos to 0.17--0.27.
Take 2 itself, scored as a one-outcome forecast ($\bar y=1$ in \cref{m:fig:decomp}a), absorbs the full $2c$ penalty and trails every full-step model on Kernel Physics-IQ over the same 52 non-duplicate scenes.
The same reversal holds beyond Physics-IQ.
On the revised Physics-IQ Verified benchmark~\cite{physicsiqverified}, two steps still raise official scores for CogVideoX and Cosmos while lowering Kernel Physics-IQ (all detected).
On CaliBench's animated Galton board~\cite{calibench}, released WAN-2.7 drops all 29 readable balls into the central bin to win single-sample one-hot distance over Runway Gen-4.5, whereas distribution-level mnTV and Brier scores rank Runway higher.
Across pixels, motion masks and discrete bins, single-sample scoring rewards shrinkage over distributional accuracy.
\section{Analysis}
\label{m:sec:sampling}\label{m:sec:ruler}
Cross-model comparisons locate the dispersion premium, yet step counts also alter visual quality.
Holding model weights fixed, we probe sampler dispersion (\cref{m:sec:node}), denoiser geometry (\cref{m:sec:ideal}), temporal history (\cref{m:sec:history}) and known-distribution optimality (\cref{m:sec:known}).

\subsection{Fixed-weight node shifts move dispersion}
\label{m:sec:node}\label{m:sec:steps}
If single-sample scoring trades distributional fidelity for shrinkage, varying dispersion at fixed weights should move Kernel Physics-IQ with no detected official difference.
We test this prediction by holding CogVideoX's weights, prompts and seeds fixed across all 66 scenes while sweeping its two-step node $t_2$ from 19 to 899 (\cref{m:fig:steps}a).
Across the six nodes, the official composite stays nearly flat while SSR and Kernel Physics-IQ rise toward later nodes.
Shifting only from default node 499 to 899 lifts Kernel Physics-IQ by a detected $+0.099$ with no detected official difference.
\mbox{Kernel} Physics-IQ still ranks the node-899 samples below 50 steps, whereas the official composite ranks them above (both detected).
Coarser two-step nodes (\cref{m:sec:ideal}) reproduce this fixed-weight split from the default node in Wan2.1 and Cosmos, with smaller detected Kernel Physics-IQ gains and no detected official difference.

Sweeping step counts with shared seeds traces how this dispersion premium fades (\cref{m:fig:steps}b).
On CogVideoX, SSR and Kernel Physics-IQ recover steadily from 1 to 50 steps, while the official composite falls from its two-step peak yet keeps a detected gain through 16 steps, where Kernel Physics-IQ already shows no detected loss.
The official gain thus outlasts any detected Kernel Physics-IQ loss, while Wan2.1 shows rising SSR with no detected difference on either score from 4 steps on.
Cosmos varies non-monotonically (\cref{m:fig:steps}c), showing that Kernel Physics-IQ does not reward larger dispersion on its own.
It shrinks at 1 and 2 steps and suffers 4-step artifacts that lower both scores.
At 8 steps, excess RMS dispersion lowers the official score with no detected Kernel Physics-IQ difference, as single-sample scoring penalizes over-dispersion too.

\subsection{Coarse steps shrink Gaussian modes}
\label{m:sec:ideal}
Sampling geometry explains how coarse discretization shrinks dispersion in an idealized setting, which our interventions test in real generators.
At zero terminal SNR~\cite{lin2024snr} (or $t=1$ in flow matching~\cite{lipman2023fm}), an exact denoiser's first evaluation returns the conditional mean $\E[x_0\mid x_T,c]=\E[x_0\mid c]$ for every seed.
One deterministic DDIM~\cite{song2021ddim} or linear Euler step therefore outputs the conditional mean for any data distribution, while infinite steps recover true samples.
On Gaussian modes of variance $\lambda$, $K$-step DDIM ($\eta=0$) and linear flow-matching Euler preserve the mean while scaling each mode's standard deviation by $s_K(\lambda)=\prod_i\cos(\theta_i-\theta_{i-1})$.
Here $\theta=\arctan\sqrt{\mathrm{SNR}}$ for mode SNR $\abar\lambda/(1-\abar)$ runs from $0$ to $\pi/2$, giving $s_K(\lambda)\le\cos^K(\pi/2K)$~\cite{ren2025,pierret2025}.
Two steps thus at least halve every mode's standard deviation, and a shrink factor $s_K$ earns a single-sample MSE gain of $\lambda(1-s_K^2)$ while leaving the $\beta=2$ kernel score invariant.
Two-step sampling $T\to t_2\to0$ acts as one-step SDEdit~\cite{meng2022sdedit} on the conditional mean with transfer
\begin{equation}
s(\lambda)=\frac{u}{1+u^2}\le\frac12,\qquad u^2=\frac{\abar_2\lambda}{1-\abar_2},
\label{m:eq:band}
\end{equation}
a band-pass centered at $\lambda^*=(1-\abar_2)/\abar_2$.
CogVideoX and Wan2.1 defaults pass coarse large-$\lambda$ modes, while Cosmos passes fine small-$\lambda$ modes.

Varying classifier-free guidance connects this idealized geometry to real networks.
Disabling guidance (CFG 1) cuts first-step latent dispersion on ten CogVideoX scenes by over 90\% and brings the one-step mean close to the 50-step mean, consistent with the zero-SNR mechanism even though nonlinear latents depart from the exact Gaussian pass-band of \cref{m:eq:band}.
Across every tested guidance scale on CogVideoX (CFG 1--9) and unguided Cosmos, two steps shrink dispersion and lower Kernel Physics-IQ.
Higher guidance narrows kernel loss faster than the premium, making 66-scene official two-step gains detected at CFG 6 and 9.
At full steps, stronger guidance likewise aligns spatial motion masks across seeds on CogVideoX and Cosmos, collecting its own dispersion premium.

\balance
\subsection{Separating history signal from shrinkage}
\label{m:sec:history}
Unlike coarse sampling, conditioning on past motion narrows uncertainty via physical information.
For any proper kernel score, adding real history $H$ alongside the switch frame $C$ cannot raise optimal risk, lowering it whenever $H$ shifts $m_{P\mid C,H}$: $R^*(C)-R^*(C,H)=\E\lVert m_{P\mid C,H}-m_{P\mid C}\rVert^2\ge0$~\cite{degroot1962,brocker2009}.
Repeating the first frame adds only a function of $C$, so the reduction vanishes.
Across the 32 history scenes at 35 steps, repeating the first frame shows no detected difference from no history on either score, whereas conditioning on the true preceding second raises Kernel Physics-IQ by 0.097.
At 2 steps, most of real history's official gain is again a dispersion premium, with no detected kernel part.

To test temporal order, we shuffle the 24 past frames in a fixed permutation on 15 of those scenes (\cref{m:fig:steps}d).
At 35 steps, shuffled history lowers Kernel Physics-IQ by 0.067 relative to real history, suggesting that the gain from real history depends on its temporal structure.
Against the repeated first frame at 2 steps, shuffled history still lifts the official score by $0.105$ through a dispersion premium, while Kernel Physics-IQ drops by $0.060$ (both detected).
Even scrambled time thus wins officially via shrinkage.\looseness=-1

\subsection{Two samples rank the true distribution first}
\label{m:sec:known}
\begin{figure}[t]
\centering
\includegraphics[width=\linewidth]{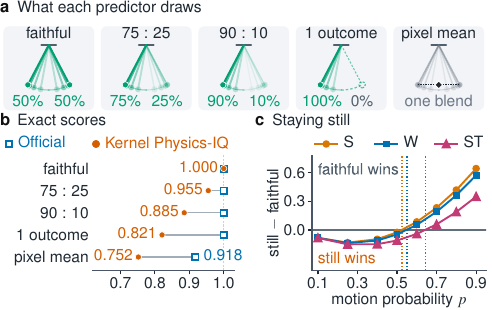}
\caption{\textbf{Tied official scores hide different futures.} (a) Predictors in a pendulum world that swings left or right with equal probability: each card shows what the predictor draws (line weight: probability; dashed: never drawn), and the pixel mean blends both swings into one video (diamond: mean position). (b) Exact scores, averaged with a ball dropped onto a peg: the raw official composite ties the first four, whereas Kernel Physics-IQ ranks the true distribution first, as it does among all 34 predictors. (c) Mask distance: staying still wins below the thresholds of \cref{m:prop:still}.}
\label{m:fig:known}
\end{figure}

Real experiments reveal two realizations of $P_c$ instead of its closed-form law.
To test whether Kernel Physics-IQ ranks the true distribution first, we construct two controlled physical worlds, a pendulum and a ball dropped onto a peg, each branched by a hidden variable (\cref{m:fig:known}a), and evaluate exact population expectations across 34 predictors.
With balanced probabilities, the official raw composite ties faithful sampling, fixed one-outcome collapse and two mixtures with wrong probabilities, whereas Kernel Physics-IQ separates all four and ranks the true distribution first among all 34 predictors (\cref{m:fig:known}b).
Among pixel scores, only the full-video RMS kernel separates truth from both one-outcome collapse and the blurry pixel mean, which single-sample MSE prefers and the $\beta=2$ MSE kernel ties.
When outcomes split 70/30, the official score rewards always outputting the majority branch over faithful sampling (1.090 versus 1.000), whereas Kernel Physics-IQ still ranks truth first.\looseness=-1

Because motion here is nearly certain, the pixel mean halves MSE while degrading all three motion masks, keeping its raw official score below faithful sampling.
Sweeping motion probability $p$ reproduces the thresholds of \cref{m:prop:still}, below which staying still beats faithful sampling (\cref{m:fig:known}c).
Clipping each seed's composite at 1, as the official code does, truncates draws that hit the recorded branch, so the pixel mean then scores highest of all 34 predictors, which is why Kernel Physics-IQ stays unclipped (\cref{m:sec:kpiq}).
A second sample realigns optimality with truth.
\section{Conclusion}
\label{m:sec:discussion}

Why can fewer denoising steps score higher on physics benchmarks?
What one sample cannot see is the model's own dispersion, yet half of it counts toward the single-sample distance.
Shrinking it thus collects a dispersion premium.
In pixels, blur collects the premium, like the storm forecast down the middle; on motion masks, standing still does.
Two-step blur and early stillness are not flaws the official score overlooks.
They are what it pays for.
A calibrated model, in turn, pays for the futures it keeps open: in expectation it lands as far from the recording as a second take does, and a deterministic forecast can land closer.
Physics-IQ already films each scene twice to gauge how much nature varies.
Kernel Physics-IQ draws two samples per condition for the same reason, much as weather forecasting scores its ensembles with the CRPS.
In expectation, a calibrated model then no longer pays for the futures it keeps open, and fewer steps earn no dispersion premium.
After all, keeping possible futures open is what a world model is for.
{
    \small
    \clearpage\nobalance
\bibliographystyle{references}
    \bibliography{main}
}

\end{document}